\pdfoutput=1

\documentclass[11pt]{article}

\usepackage[]{ACL2023}
\usepackage{graphicx}
\usepackage{subcaption}
\usepackage{booktabs,tabularx}

\usepackage{comment}

\usepackage{hyperref}
\usepackage{amsmath}
\usepackage{graphicx}

\usepackage{times}
\usepackage{latexsym}
\usepackage{adjustbox}
\usepackage{lipsum}

\usepackage[T1]{fontenc}

\usepackage[utf8]{inputenc}

\usepackage{microtype}

\makeatletter
\def\thanks#1{\protected@xdef\@thanks{\@thanks
        \protect\footnotetext{#1}}}
\makeatother

\title{Trillion Dollar Words: A New Financial Dataset, Task \& Market Analysis$^*$}

\author{\hypersetup{linkcolor=black} Agam Shah\;, Suvan Paturi\;, Sudheer Chava\\
Georgia Institute of Technology
\thanks{* Accepted for presentation at the ACL 2023 (main)}
\thanks{Correspondence to Agam Shah \textcolor{darkblue}{{\{\href{mailto:ashah482@gatech.edu}{ashah482@gatech.edu}\}}}}}

\begin{document}
\maketitle
\begin{abstract}
 Monetary policy pronouncements by Federal Open Market Committee (FOMC) are a major driver of financial market returns. We construct the largest tokenized and annotated dataset of FOMC speeches, meeting minutes, and press conference transcripts in order to understand how monetary policy influences financial markets. In this study, we develop a novel task of hawkish-dovish classification and benchmark various pre-trained language models on the proposed dataset. Using the best-performing model (RoBERTa-large), we construct a measure of monetary policy stance for the FOMC document release days. To evaluate the constructed measure, we study its impact on the treasury market, stock market, and macroeconomic indicators. Our dataset, models, and code are publicly available on Huggingface and GitHub under CC BY-NC 4.0 license\footnote{The fine-tuned model and data are available on the \href{https://huggingface.co/gtfintechlab/FOMC-RoBERTa}{Huggingface}. The code is available on \href{https://github.com/gtfintechlab/fomc-hawkish-dovish}{FinTech Lab GitHub}.}.

\end{abstract}

\section{Introduction}

On August 26th, 2022, FOMC Chair Jerome H. Powell gave an 8-minute long speech at Jackson Hole which immediately resulted in an almost \$3 Trillion USD decline in U.S. equity market value that day. The speech was followed by more than \$6 Trillion USD loss in equity valuation over the next 3 days. Drastic market shifts to the Fed's pronouncements indicate just how important the FOMC communications have become and highlight the need for a model which can capture the policy stance from Fed-related text. 

The Federal Open Market Committee (FOMC) is a federal organization responsible for controlling U.S.'s open market operations and setting interest rates. It tries to achieve its two main objectives of price stability and maximum employment by controlling the money supply in the market. Given the market condition (employment rate and inflation), the Fed either increases (dovish), decreases (hawkish), or maintains the money supply\footnote{Fed increases the money supply by lowering interest rates and decreases the money supply by increasing interest rates or by other means necessary. More detail on this can be found in the annotation guide. } (neutral). To understand the influence the FOMC has on the different financial markets, we need to extract its monetary policy stance and the corresponding magnitude from official communications. 

Utilizing the traditional sentiment analysis model, which classifies text into positive vs negative, one can't extract policy stance. A sentence that has the word "increase" could either be dovish or hawkish without a clear negative connotation. For example, the word "increase" with the word "employment" means the economy is doing well, but the word "increase" with the word "inflation" is negative for the economy. Current SOTA finance domain-specific language models \citep{orig_finbert, shah-etal-2022-flang} trained for sentiment analysis find both cases to be positive, which is inaccurate. The performance analysis for FinBERT \citep{orig_finbert} model is provided in Appendix \ref{ap:finbert_senti}. This problem creates a need to develop a new task for hawkish vs dovish classification accompanied by high-quality annotated data. 

Given the lack of annotated data, computational linguistic work related to FOMC text in the literature \citep{rozkrut2007quest, zirn-etal-2015-lost, hansen2016shocking, rohlfs-etal-2016-effects, hansen2018transparency, nakamura2018high, cieslak2019stock, schmeling2019does, tsukioka2020tone, ehrmann2020starting, frunza-2020-information, gorodnichenko2021voice, matsui-etal-2021-using, 10.1145/3503161.3548380} so far has been limited to unsupervised and rule-based models. These rule-based models don't perform well on the hawkish-dovish classification task, which we will use as a baseline in performance analysis. Additionally, we conduct a benchmark of the zero-shot ChatGPT model using the annotated dataset to gain insights into the significance of fine-tuning on such data. In this work, we collect text data (speech transcripts, press conference transcripts, and meeting minutes) from the FOMC over the period 1996-2022 and annotate a sample of each data type.

We not only create new datasets and tackle the task of building a hawkish-dovish classifier, but also test the performance of various models starting from rule-based to fine-tuned large PLMs. As sentences presented in FOMC text sometimes have two sub-sentences that have counterfactual information to tone down the stance, we employ a simple sentence-splitting scheme as well. We also construct the aggregate monetary policy stance and show its validity by looking at its performance in predicting various financial market variables.

Through our work, we contribute to the literature in the following way:
\begin{itemize}
    \item We show that the traditional (rule-based) approach practiced in finance and economic literature is a rudimentary way to measure monetary policy stance from the text document. 
    \item We introduce a new task to classify sentences into hawkish vs dovish as opposed to positive vs negative sentence classification for monetary policy text.
    \item We build comprehensive, clean, tokenized, and annotated open-source datasets for FOMC meeting minutes, press conferences, and speeches with detailed meta information. 
    \item We develop an aggregate monetary policy stance measure and validate its performance in predicting various economic and financial indicators. 
\end{itemize}

\section{Related Work}

\paragraph{NLP in Finance} Over the last decade behind the evolution of NLP, there has been a growing literature on the applications of NLP techniques in Finance \citep{loughran2011liability, sohangir2018big, xing2018natural, chava2022measuring}. The majority of the research takes advantage of news articles \citep{vargas2017deep, caldara2022measuring}, SEC filings \citep{loughran2011liability, chava2016december, alanis2022benchmarking}, or earnings conference calls \citep{bowen2002conference, bushee2003open, chava2019buzzwords, li2020maec}. Development of finance domain-specific language models \citep{orig_finbert, finbert, liu2020finbert} have pushed the current benchmarks further. Recent work of \citet{shah-etal-2022-flang} proposes a set of heterogeneous benchmarks for the financial domain and shows SOTA performance using their proposed language model, but it doesn't include macroeconomics-based tasks.  

\paragraph{FOMC and Text Analysis} A study on communications from the central banks of the Czech Republic, Hungary, and Poland by \citet{rozkrut2007quest} suggests that words from central banks affect the market but the effect varies based on communication style. Other various studies \citep{tobback2017between, hansen2018transparency, nakamura2018high, cieslak2019stock, schmeling2019does, tsukioka2020tone, ehrmann2020starting, bennani2020does, gorodnichenko2021voice} also point to a similar conclusion that the communication from the central banks moves the market, but they don't leverage the power of the transformer-based model available at their disposal. 

Many articles in the literature use LDA to analyze various texts released by Fed. \citet{rohlfs-etal-2016-effects} uses LDA on the FOMC meeting statements to predict the fed fund rate and long-term treasury rate. \citet{hansen2016shocking} use an LDA-based topic modeling on FOMC-released text to understand how forward guidance affects the market and economic variables. In their study, they only used statements released post-meeting and suggest that the use of meeting minutes and speeches may offer greater insight. \citet{jegadeesh2017deciphering} also uses LDA to analyze meeting minutes. They suggest that even though meeting minutes are released a few weeks after the actual meeting, the minutes still carry pertinent market-moving information. 

In recent work by \citet{10.1145/3503161.3548380}, they created a multimodal dataset (MONOPOLY) from video press conferences for multimodal financial forecasting. The MONOPOLY dataset is comprehensive and not only covers text but also utilizes audio and video features. Yet, it misses two critical economic downturn periods of the last two decades: The DotCom Bubble Burst of 2000-2002 and the Global Financial Crises of 2007-2008.  \citet{matsui-etal-2021-using} used word embedding to extract semantic changes in the monetary policy documents. \citet{zirn-etal-2015-lost} used the graph clustering method to generate the hawkish-dovish stance of monetary policy due to the dearth of annotated data. \citet{frunza-2020-information} developed an unsupervised methodology to extract various information from FOMC post-meeting statements.

\section{Dataset}
\subsection{FOMC Data}

The datasets we build are composed of three different types of data: meeting minutes, press conference transcripts, and speeches from the FOMC.
Meeting minutes are defined as reports derived from the eight annually scheduled meetings of the FOMC. Press conference transcripts, meanwhile, are transcripts of the prepared remarks, followed by the Q\&A session between the Federal Reserve chair and press reporters. Lastly, speeches were defined as any talk given by a Federal Reserve official. We limit our datasets to an end release date of October 15th, 2022, and attempt to collect as far back as possible for each category prior to this date.

The meeting minutes and speeches spanned from a release period of January 1st, 1996 to October 15th, 2022. Press conferences are a more recent phenomenon and the data aggregated stretched from April 27th, 2011 to October 15th, 2022. We obtained the data by leveraging BeautifulSoup, Selenium, and manual downloading from \url{http://www.federalreserve.gov/}. Regex tools were used to clean the data, which was stored in CSV or Excel format for processing. Sentence tokenization, using the library NLTK \cite{bird2009natural} was done and datasets for each data category were initialized.

\paragraph{FOMC Raw Text Data} The overview of our initial raw text dataset is presented in Panel A of Table~\ref{tb:raw_text_data_info}.
Initial observations show that meeting minutes and speeches composed the bulk of our data, due to the recency of press conference transcripts. In addition, we also isolated only sentences where the speaker is designated as the Federal Reserve chair and the sentence was not a question in press conference transcripts, so this also served to reduce the data size.
Across all forms of data, we had higher average words per sentence than the typical English language sentence, which averages 15 to 20 words \cite{cutts2020oxford}. 

Our initial raw text data encompassed decades worth of crucial FOMC statements, however, a plethora of noise persisted in the data.
Unrelated sentences riddled the datasets and a filter was needed to isolate key sentences relevant to changes in the federal reserve's monetary policy stance. In addition, the number of sentences in the raw dataset was too vast to manually label, so a sampling procedure was needed.

\paragraph{Data \& Title Filtration}
As a result of data noise, a dictionary filter was developed to isolate sentences that would prove to be meaningful and allow us to determine monetary policy stance. The criteria for the filter was based on the dictionary developed by \citet{gorodnichenko2021voice}. Any sentence that contained an instance of the words outlined in panel A1 or B1 in Table \ref{tb:rule-based}
would be kept, while anything else would be filtered out. The sentences kept were considered "target" sentences or textual data that we consider pertinent and later used to sample from and annotate. 

\begin{table}[ht]
\centering
\footnotesize
\begin{tabular}{p{0.25\textwidth}|p{0.2\textwidth}}
\hline
\textbf{Panel A1} & \textbf{Panel B1}\\
\hline
inflation expectation, interest rate, bank rate, fund rate, price, economic activity, inflation, employment
& 
unemployment, growth, exchange rate, productivity, deficit, demand, job market, monetary policy\\
\hline
\textbf{Panel A2} & \textbf{Panel B2}\\
\hline
anchor, cut, subdue, decline, decrease, reduce, low, drop, fall, fell, decelerate, slow, pause, pausing, stable, non-accelerating, downward, tighten & 
ease, easing, rise, rising, increase, expand, improve, strong, upward, raise, high, rapid \\
\hline
\textbf{Panel C} \\
\hline
weren't, were not, wasn't, was not, did not, didn't, do not, don't, will not, won't \\
\hline

\end{tabular}
\caption{Rule-based dictionary used by \citeauthor{gorodnichenko2021voice}}
\label{tb:rule-based}
\end{table}

Our dictionary filter was also applied to speech data. Speech data was the largest dataset derived from web scraping, however, speeches contained the most noise, owing to many non-monetary policy speeches.
Unlike the meeting minutes and press conference transcripts, speech data was accompanied with a title, so to isolate only relevant FOMC speeches to sample from, we applied the dictionary filter discussed in Table \ref{tb:rule-based} onto the title of each speech. We justify this procedure in Table \ref{tb:speech-filter} as this methodology results in the greatest "target" sentence per file. Overall, the filtration process isolated relevant files and "target" sentences in our raw data and set the stage for later sampling. The filter's impact on the raw data is presented in Panel B of Table \ref{tb:raw_text_data_info}.

\begin{table*}
\centering
\footnotesize
\begin{tabular}{ccccc}
\hline
\textbf{Type} & \textbf{\# Files} &\textbf{\# Sentences} & \textbf{\# Target Sentences} & \textbf{\# Target Sentences per File} \\
\hline
All Speech Titles & 1,026 & 108,463 & 27,221 & 26.53\\
Non-Filtered Speech Titles & 825 & 84,833 & 14,756 & 17.89 \\
Filtered Speech Titles & 201 & 23,630 & 12,465 & \textbf{62.01}\\
\hline 
\end{tabular}
\caption{Details on the speech title filter methodology}
\label{tb:speech-filter}
\end{table*}

\begin{table*}
\centering
\footnotesize
\begin{tabular}{cccccc}
\hline
\textbf{Event} & \textbf{Years} & \textbf{\# Files} & \textbf{\# Sentences} & \textbf{\# Words}  & \textbf{Avg. Words in Sentence} \\
\hline
\multicolumn{6}{c}{Panel A: Pre-Filter}\\
\hline
Meeting Minutes & 1996 - 2022 & 214 & 44,923 & 1,346,674  & 29.98 \\
Meeting Press Conferences & 2011-2022 & 63 & 19,068 & 468,941  & 24.59 \\
Speeches & 1996-2022 & 1,026 &  108,463 & 3,222,285  & 29.71 \\
\hline
\multicolumn{6}{c}{Panel B: Post-Filter}\\
\hline
Meeting Minutes & 1996 - 2022 & 214 & 20,618 & 692,759  & 33.60 \\
Meeting Press Conferences & 2011-2022 & 63 & 5,086 & 160,574  & 31.57 \\
Speeches & 1996-2022 & 201 & 12,465 & 447,974  & 37.62 \\
\hline
\end{tabular}
\caption{Details on the text data covered from FOMC}
\label{tb:raw_text_data_info}
\end{table*}


\paragraph{Sampling and Manual Annotation} As our data was unlabeled, our analysis necessitated the usage of manual labeling. To efficiently develop a manually labeled dataset, sampling was required. Our sampling procedure was to extract 5 random sentences and compile a larger data set. If fewer than 5 sentences were present in the file, all sentences were added. This sampling procedure resulted in a 1,070-sentence Meeting Minutes dataset, a 315-sentence Press Conference dataset, and a 994-sentence Speech dataset. For the labeling process, sentences were categorized into three classes (0: Dovish, 1: Hawkish, and 2: Neutral). We annotate each category of the data as a model trained on various categories as a model trained on the same category of data does not perform optimally. We provide evidence for this claim in Appendix \ref{ap:transfer_learning}. 


Dovish sentences were any sentence that indicates future monetary policy easing. Hawkish sentences were any sentence that would indicate a future monetary policy tightening. Meanwhile, neutral sentences were those with mixed sentiment, indicating no change in the monetary policy, or those that were not directly related to monetary policy stance.

The labeling was conducted by two different annotators and done independently to reduce potential labeling bias. Each annotator's labeling was compared against each other and validated to ensure the consistency of the labeling results. The detail on the annotation agreement is provided in Appendix \ref{sec:agreement_ann}. The labeling was conducted according to a predefined annotation guide, which is provided in Appendix \ref{sec:appendix_manual_ann}. The guide is broken down into key sections such as economic status, dollar value change, energy/house prices, future expectations, etc.

\paragraph{Sentence Splitting} A common occurrence in the labeling process was the existence of intentional mixed tone. The Federal Reserve by purpose serves to maintain financial/economic stability and any statement they make is projected in a moderating manner to reduce the chance of excess market reaction. As a result, the Fed is known to project a stance but often accompanies this with a moderating statement that serves as a counterweight to the original stance. This produces a greater occurrence of neutral sentences. To address this possibility, we instituted sentence splitting to separate the differing stances. Initially, we implemented the lexicon-based package SentiBigNomics \citep{SentiBigNomics} for sentence splitting, but it resulted in poor performance, causing us to pivot our approach. We developed a custom sentence-splitting method based on keywords. In Fed statements, the counter-statements are produced after a connective contrasting word. We carried sentence splits at the presence of the following keywords in a given statement: "but", "however", "even though", "although", "while", ";". A sentence split was valid if each split segment contained a key word present in Table \ref{tb:rule-based}. Statistics on the dataset before and after splitting are provided in Table \ref{tb:before_after_split}. 

\begin{table}[h]
\centering
\footnotesize
\begin{tabular}{ccc}
\hline
\textbf{Event} & \textbf{Before split} & \textbf{After split} \\
\hline
Meeting Minutes & 1,070 & 1,132 \\
Meeting Press Conferences & 315 & 322 \\
Speeches & 994 & 1,026  \\
\hline
Total & 2,379 & 2,480\\
\hline
\end{tabular}
\caption{Number of sentences in the labeled dataset before and after splitting for each event. }
\label{tb:before_after_split}
\end{table}

\subsection{Economic Data}
\paragraph{CPI and PPI} We collect Consumer Price Index (CPI) data, and Producer Price Index (PPI) data from FRED\footnote{\url{https://fred.stlouisfed.org}}. The data is available at the monthly frequency for the first day of each month. Throughout the paper, we use percentage change from last year as CPI and PPI inflation measures. 

\paragraph{US Treasury} We collect US treasury yield data for different maturities from the U.S. Department of the Treasury\footnote{\url{https://home.treasury.gov}}. It provides a daily yield of bonds for various maturities.  

\paragraph{QQQ Index} We collect the adjusted closing index price of QQQ from Yahoo Finance\footnote{\url{https://finance.yahoo.com/quote/QQQ/history?p=QQQ}}. It contains daily QQQ index data since March 9, 1999.

\section{Models}
\subsection{Rule-Based}
In financial literature, rule-based classification has been the norm. Many of these rule-based systems work by classifying based on the presence of a combination of keywords. \citet{gorodnichenko2021voice} in particular highlighted the effectiveness of this approach by classifying sentences as dovish or hawkish based on the combination of financial-related nouns and verbs in set panels in a given sentence. We have applied \citeauthor{gorodnichenko2021voice}'s financial word dictionary rule-based approach to our developed datasets. In Table \ref{tb:rule-based}, a sentence is considered dovish if it contains words present in panels A1 and A2 or B1 and B2. Otherwise, if it contains words present in A1 and B2 or A2 and B1 are considered hawkish. If a given sentence contains a word from panel C we reverse our initial classification, so dovish becomes hawkish and vice versa.
We aim to capture and measure the effectiveness of the rule-based approach against our dataset to provide a benchmark against the deep learning models we apply later. We apply this rule-based approach on testing datasets that we derive from each dataset on an 80:20 training-test split.

\subsection{LSTM \& Bi-LSTM}
Long short-term memory (LSTM) is a recurrent neural network structure utilized for classification problems. The Bi-LSTM is a variation of an LSTM, which takes input bidirectionally. We apply both an LSTM and a Bi-LSTM to our developed datasets to gauge the effectiveness of RNNs in monetary stance classification. 
We instituted an 80:20 training-validation split to derive our initial training and validation datasets.
A vocabulary was developed for both models against the training dataset for the purpose of vectorization. The encoding process worked by first initializing a tokenizer that eliminated all punctuation, normalized all sentences to lowercase, and splits sentences into word tokens. We limit the vocabulary size to 2,000 and any words outside the vocabulary were replaced with a placeholder token. A vocabulary size of 2,000 covers more than 99\% of words in MM and PC text and covers around 91\% in SP text. The lower coverage for speeches is due to the wide variety of miscellaneous topics outside of the scope of monetary policy. 
Our vocabulary allowed us to convert each sentence into a word vector by mapping each word to a corresponding numerical value present in the dictionary. Each word vector size was set to the length of the longest sentence present in the training dataset, and padding was done to meet the required vector size. We applied this encoding process to the training, testing, and validation datasets. 
Upon the complexion of vectorization, the word vectors were passed into our single-layer LSTM (32,379 parameters) and single-layer Bi-LSTM (32,735 parameters) models. Masking was also configured to true to ignored padded data and dropout was added to reduce potential over-fitting. We ran each model at varying epochs (10, 20, 30) and batch sizes (4, 8, 16, 32). Implementation of models was done using Tensorflow \citep{tensorflow} on an NVIDIA RTX A6000 GPU. 

\subsection{PLMs}
To set a benchmark, we include a range of small and large transformer-based models in our study. For small models, we use BERT \citep{bert}, FinBERT \citep{finbert}, FLANG-BERT \citep{shah-etal-2022-flang}, FLANG-RoBERTa \citep{shah-etal-2022-flang}, and RoBERTa \citep{roberta}. In the large model category, we include BERT-large \citep{bert} and RoBERTa-large \citep{roberta}. We do not perform any pre-training on these models before employing them for fine-tuning to avoid overfitting on FOMC text. For each model, we find best hyper-parameters by performing a grid search on four different learning rates (1e-4, 1e-5, 1e-6, 1e-7) and four different batch sizes (32, 16, 8, 4). We conduct all experiments using PyTorch \citep{pytorch} on NVIDIA RTX A6000 GPU. Each model was initialized with the pre-trained version on the Transformers library of Huggingface~\citep{huggingface}. 

\subsection{ChatGPT}
In order to provide the performance benchmark of the current SOTA generative LLM, we measure the zero-shot performance of ChatGPT. We use the "gpt-3.5-turbo" model with 1000 max tokens for output, and a 0.0 temperature value. All the API calls were made on either May 3rd, 2023, or May 4th, 2023. We use the following zero-shot prompt: 

"Discard all the previous instructions. Behave like you are an expert sentence classifier. Classify the following sentence from FOMC into `HAWKISH', `DOVISH', or `NEUTRAL' class. Label `HAWKISH' if it is corresponding to tightening of the monetary policy, `DOVISH' if it is corresponding to easing of the monetary policy, or `NEUTRAL' if the stance is neutral. Provide the label in the first line and provide a short explanation in the second line. The sentence: \{sentence\}"

\begin{table*}[ht]
\centering
\footnotesize
\begin{tabular}{lcccccccc}
  \hline
  \textbf{Model} & \textbf{MM} & \textbf{MM-S} & \textbf{PC} & \textbf{PC-S} & \textbf{SP} & \textbf{SP-S} & \textbf{Combined} & \textbf{Combined-S} \\ 
  \hline
  Rule-Based & 0.5216 & 0.5200 & 0.4927 &  0.5114 & 0.5449 & 0.5388 & 0.4966 & 0.5165 \\ 
 & (0.0432) & (0.0298) & (0.0387) & (0.0661) & (0.0286)
 & (0.0038) & (0.0041) & (0.0064)\\
  \hline
  LSTM & 0.4982 & 0.5192 & 0.3373 & 0.2877 & 0.4828 & 0.5352 & 0.4917 & 0.5247\\ 
  & (0.0396) & (0.0315) & (0.0172) & (0.1287) & (0.0564) & (0.045) & (0.027) & (0.0251)\\
  Bi-LSTM & 0.4906 & 0.5175 & 0.3321 & 0.3186 & 0.5296 & 0.5106 & 0.5387 & 0.5089\\ 
  & (0.0679 & (0.0310) & (0.0876) & (0.0853) & (0.0334) & (0.0712) & (0.0213) & (0.0608)\\
  \hline
  BERT-base-uncased & 0.5889 & 0.6115 & 0.4676 &  0.5227 & 0.6151 & 0.6007 & 0.6310 & 0.6360 \\ 
       & (0.0525) & (0.0419) & (0.0883) &  (0.0472) & (0.0201) & (0.0524) & (0.0086) & (0.0225) \\ 
  FinBERT-base & 0.6173 & 0.6486 & 0.4631 & 0.5452 & 0.6595 & 0.6291 & 0.6325 & 0.6304 \\ 
  -uncased     & (0.0413) & (0.0126) & (0.0368) & (0.0587) & (0.0053) & (0.0300) & (0.0172) & (0.0217) \\ 
  FLANG-BERT-base & 0.6334 & 0.6360 & 0.4647 & 0.5132 & 0.6412 & 0.6355 & 0.6307 & 0.6443 \\ 
  -uncased        & (0.0258) & (0.0201) & (0.0726) & (0.0830) & (0.0308) & (0.0489) & (0.0192) & (0.0117) \\
  FLANG-RoBERTa & 0.6446 & 0.6854 & 0.4995 &  0.4666 & 0.6745 & 0.5561 & 0.6618 & 0.6348 \\ 
  -base      & (0.0185) & (0.0035) & (0.0413) &  (0.0732) & (0.0141) & (0.1559) & (0.0065) & (0.0021) \\ 
  RoBERTa-base & 0.6741 & 0.6752 & 0.5371 & \textbf{0.5527} & 0.6885 & 0.6725 & 0.6755 & 0.6981 \\ 
          & (0.0096) & (0.0119) & (0.0102) & (0.0589) & (0.0261) & (0.0147) & (0.0267) & (0.0097) \\ 
  \hline
  BERT-large-uncased & 0.6843 & 0.6560 & 0.4013 &  0.5231 & 0.6208 & 0.6434 & 0.6579 & 0.6619 \\ 
             & (0.0486) & (0.0058) & (0.0752) &  (0.0275) & (0.0581) & (0.0076) & (0.0195) & (0.0123) \\ 
  RoBERTa-large & \textbf{0.7150} & \textbf{0.7128} & \textbf{0.5517} &  0.5346 & \textbf{0.7169} & \textbf{0.7049} & \textbf{0.7171} & \textbf{0.7113} \\
                & (0.0096) & (0.0137) & (0.0526) &  0.0580 & (0.0125) & (0.0298) & (0.0164) & (0.0106) \\
   \hline
   ChatGPT-3.5-Turbo & 0.5671 & 0.5997 & 0.4869 &  0.5222 & 0.6446 & 0.6120 & 0.5872 & 0.5868 \\
                     & (0.0359) & (0.0186) & (0.0370) &  (0.0882) & (0.0377) & (0.0153) & (0.0199) & (0.0131) \\
    \hline
\end{tabular}

\caption{Here MM indicates that the annotated dataset on meeting minutes is used for training and testing hawkish vs dovish task. Similarly, PC stands for press conference data, SP stands for speech data, and Combined is combined data of MM, SP, and PC. *-S indicates the version of the dataset after splitting sentences and reannotation. All  values are F1 scores. An average of 3 seeds was used for all models. The standard deviation of F1 scores is reported in parentheses on the next line. ChatGPT and rule-based models are tested as zero-shot while all other models are fine-tuned with training data. }
\label{tb:master_accuracy}
\end{table*}

\begin{figure*}[ht]
\centering
   \begin{subfigure}{0.9\linewidth}
   \centering
   \includegraphics[width=\linewidth]{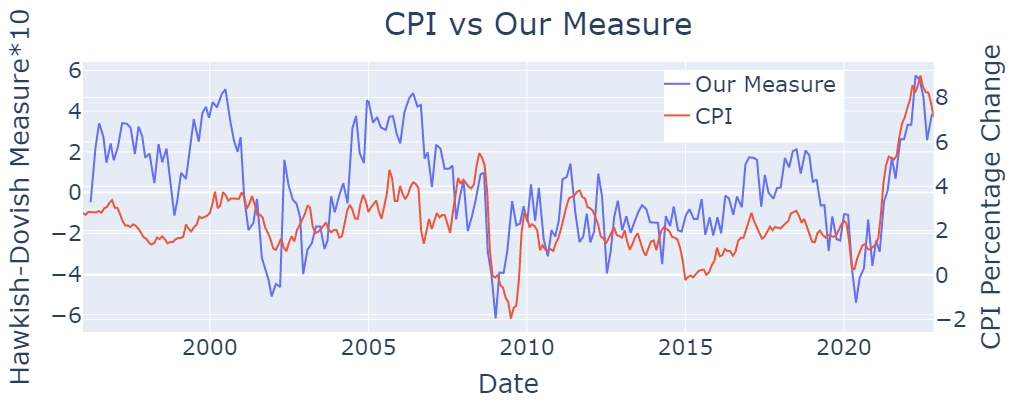}
   \caption{}
   \label{fig:CPI_measure} 
\end{subfigure}
\hfill
\\[\baselineskip]
\begin{subfigure}{0.9\linewidth}
   \centering
   \includegraphics[width=\linewidth]{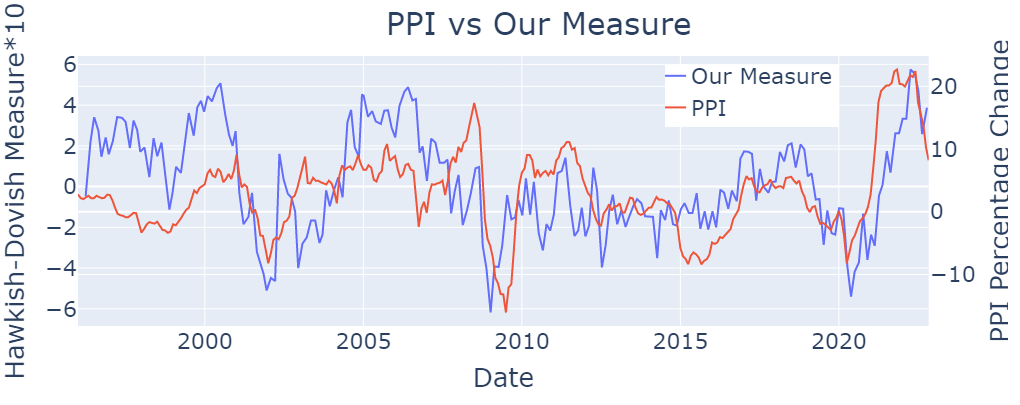}
   \caption{}
   \label{fig:PPI_measure}
\end{subfigure}
\centering
\caption{(a) Our measure on meeting release date and 1-year change in CPI data on the first day of each month (b) Our measure on meeting release date and 1-year change in PPI data on the first day of each month}
\label{fig:CPI_PPI_measure}
\end{figure*}

\begin{table*}[ht]
\centering
\footnotesize
\begin{tabular}{ccccc}
\hline
\textbf{Chair (Years)} & \textbf{Correlation (CPI)} & \textbf{Correlation (PPI)} & \textbf{Avg. Delay (days)}\\
\hline
Full Sample (1996-2022) & 0.54(1.2e-17) & 0.45(4.1e-12) & 29.78\\
\hline
Greenspan (1996-2006) & 0.46(2.0e-5) & 0.42(8.4e-5) & 44.15\\
Bernanke (2006-2014) & 0.51(1.9e-5) & 0.40(1.0e-3) & 20.97\\
Yellen (2014-2018) & 0.55(1.2e-3) & 0.57(6.2e-4) & 21.00\\
Powell (2018-2022) & 0.81(8.4e-10) & 0.71(9.9e-7) & 21.13\\
\hline
Speeches (1996-2022) & 0.58(2.6e-19) & 0.39(1.2e-8) & 0\\
\hline
Press Conferences (2011-2022) & 0.78(6.3e-14) & 0.68(8.6e-10) & 0\\
\hline
\end{tabular}
\caption{Correlation of immediate next CPI and PPI data with our measure. All values are statistically significant. The value in parentheses represents the corresponding p-value. CPI and PPI are the percentage change from last year. }
\label{tb:CPI_PPI_corr}
\end{table*}

\section{Results and Analysis}
In this section, we evaluate and benchmark different NLP models on the hawkish vs dovish classification task that we created. For all models and datasets, we used training and testing data based on an 80:20 split. Upon this split, we institute another 80:20 split on the training data to generate our final training and validation data. We use the best-performing model (RoBERTa) to generate a document (event) level measure of hawkish tone. We then validate the generated measure by looking at its relation with the inflation indicators and the US treasury. We also look at the performance of a simple trading strategy based on the generated measure. 

\subsection{Model Performance}
We ran all models listed in the previous section on three different categories and combined data. For each dataset, we train and test each model on both the before-split and after-split versions of sentences. For each model, we use three different seeds (5768, 78516, 944601) and calculate the average weighted F1 scores. The results for best hyper-parameters are listed in Table~\ref{tb:master_accuracy}.

\paragraph{Rule-Based}
As expected the rule-based model doesn't perform very well. The rule-based approach optimizes the time needed for classification, but sacrifices the nuance of complex sentences, which necessitate context. It gives an F1 score of around 0.5 for nearly all datasets. The method sets a good baseline for the dataset as it's still widely used in econ literature. 

\paragraph{LSTM \& Bi-LSTM}
Although the LSTM and Bi-LSTM models are able to utilize greater context for classification, they did not perform significantly better than the initial rule-based approach. As seen across all data categories, the RNN models performed marginally the same. The LSTM and Bi-LSTM performances largely differed between the data categories. They performed worst when applied to the press conference datasets, a discrepancy caused by the small size of the dataset. In fact, in the smaller press conference datasets, the rule-based
performed better than the expected RNN approach. Unlike rule-based approaches, neural network classification requires a large database to train from to improve accuracy.
Concurrently, the recurrent neural networks worked best when applied to the meeting minutes and speech datasets. When compared against all data categories, the Bi-LSTM did not perform significantly better than the LSTM itself. The RNNs are effective in sentence classification, yet their limited success with FOMC sentences demonstrates the need for a transformer-based model.

\paragraph{PLMs} 
Finetuned PLMs outperform rule-based model and LSTM models by a significant margin. In base size, RoBERTa-base outperforms all other models on all datasets except after-split meeting minutes data (MM-S). On PC, FLANG-RoBERTa performs best. A future study using ablation states of models to understand why the finance domain-specific language models don't outperform RoBERTa and how they can be improved could be fruitful. In large category and overall, RoBERTa large provide the best performance across all categories except PC-S.  

We note that sentence splitting does help improve performance for meeting minutes and press conference data, but it doesn't help with speech data. Also, on average improvement from sentence splitting is higher with the base models compared to large models. The goal of sentence splitting is to not improve the performance of the classification task but to better measure, the document-level monetary policy stance constructed in the next section. In order to make sure that there is no look-ahead bias in our performance, we perform a robustness check in Appendix \ref{ap:robust}. 

\paragraph{ChatGPT} 
Zero-shot ChatGPT outperforms both rule-based and fine-tuned RNN-based (LSTM \& Bi-LSTM) models. We note that the ChatGPT can't be considered a good baseline as it has many issues highlighted by \citet{rogers-etal-2023-closed}. ChatGPT model with zero-shot underperforms fine-tuned PLMs across all datasets. The finding here is in line with the survey done by \citet{pikuliak_chatgpt_survey}, which finds that zero-shot ChatGPT fails to outperform fine-tuned models on more than 77\% of NLP tasks.

\subsection{Hawkish Measure Construction}
We use the RoBERTa-large model finetuned on the combined data to label all the filtered sentences in the meeting minutes, speeches, and press conferences. We then use labeled sentences in each document to generate a document-level measure of hawkishness for document $i$ using the following formula:

\begin{equation*}
    Measure_{i} = \frac{\#Hawkish_{i} - \#Dovish_{i}}{\#Total_{i}}
\end{equation*}

where $Measure_{i}$ is document level measure, $\#Hawkish_{i}$ is number of hawkish sentences in document $i$, $\#Dovish_{i}$ is number of dovish sentences in document $i$, and $\#Total_{i}$ is the total number of filtered sentences.

\subsection{Market Analysis}

\paragraph{Our Measure with CPI and PPI} To understand how quick the Fed is in reacting to inflation or deflation we use monthly CPI and PPI data and overlay our measure. As observed in Figure~\ref{fig:CPI_PPI_measure}, our measure based on meeting minutes captures both the inflation and deflation period pretty well. It also shows that when Fed reacts quickly (2001 and 2008) it controls inflation and deflation better.

We also look at the correlation of our measure with the CPI and PPI percentage change. As reported in Table~\ref{tb:CPI_PPI_corr}, for all three data classes we find a statistically significant positive correlation. We also observe that the correlation increases over time as Fed is communicating its policy stance better to the public in recent years. As part of better communication, the Fed has started hosting press conferences at every alternate meeting starting in 2011 and every meeting starting in 2019. We refer readers to \citet{coibion2022monetary} for a detailed discussion on Fed communication shift over time.

\begin{table}[ht]
\centering
\footnotesize
\begin{tabular}{ccc}
\hline
\textbf{Maturity} & \textbf{Constant ($\alpha$)} & \textbf{Beta ($\beta$)} \\
\hline
\multicolumn{3}{c}{Panel A: Meeting Minutes (1996-2022)}\\
\hline
3 Month & 1.94*** & 4.91*** \\
1 Year & 2.17*** & 5.23*** \\
10 Year & 3.54*** & 3.12*** \\
\hline
\multicolumn{3}{c}{Panel B: Speeches (1996-2022)}\\
\hline
3 Month & 2.64*** & 1.69** \\
1 Year & 2.82*** & 2.21*** \\
10 Year & 3.81*** & 1.79*** \\
\hline
\multicolumn{3}{c}{Panel C: Press Conf. (2011-2022)}\\
\hline
3 Month & 0.76*** & 1.27** \\
1 Year & 0.98*** & 1.97*** \\
10 Year & 2.10*** & 1.10** \\
\hline
\end{tabular}
\caption{Regression of Treasury yield with different maturity on our measure. All values are statistically significant.  *, **, and *** indicate significance at the 10\%, 5\%, and 1\% levels, respectively. }
\label{tb:treasury_reg}
\end{table}

\paragraph{US Treasury Market} is highly sensitive to monetary policy changes. We validate the power of our measure in estimating treasury yield by running the linear regression provided in the Eq~\ref{eq:treasury_reg}. We run the regression for three different maturities (3 months, 1 year, and 10 years) using three time-series measures generated from meeting minutes, speeches, and testimonies. We report the results in Table~\ref{tb:treasury_reg}. We observe that the yield of treasury with 1-year maturity is most sensitive to monetary policy changes. All the regression yields statistically significant results which further validate the generated measure. 

\begin{equation}
    Yield_{t,T} = \alpha_{T} + \beta_{T} * Measure_{t} + \epsilon_{t,T}      
    \label{eq:treasury_reg}
\end{equation}

here $T$ indicates maturity, and $t$ indicates the date on which the document was released.

\begin{figure}[ht]
    \centering
    \includegraphics[width=\linewidth]{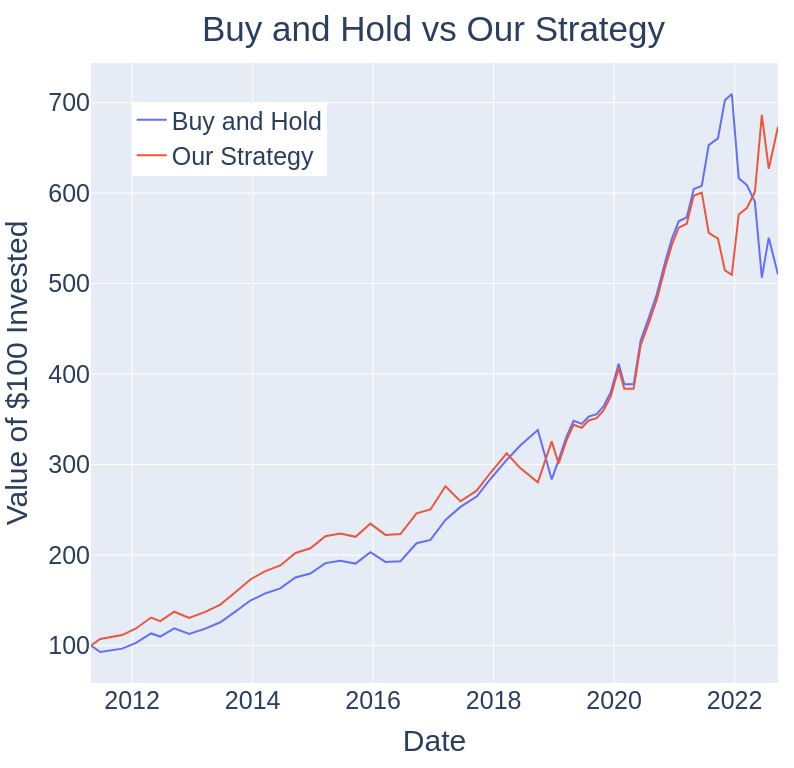}
    \caption{Value of \$100 portfolio over time for two different strategies (Buy and Hold, and Our Strategy).}
    \label{fig:trading_qqq}
\end{figure}

\paragraph{Equity Market} For a reality check, we construct a simple trading strategy based on the generated measure and compare its performance against the "Buy and Hold" strategy. In our strategy, we take a short position of the QQQ index fund when the measure is positive (hawkish) and a long QQQ position when the measure is negative (dovish). In the "Buy and Hold" strategy, the portfolio is always long QQQ. As shown in Figure~\ref{fig:trading_qqq}, our strategy provides an excess return of 163.4\% (673.29\% our strategy vs 509.89\% buy and hold) compared to the buy and hold strategy as of September 21st, 2022. Not only did our strategy outperform at the end, but it gives a better return during the majority of the period. We analyze the strategy for the period for which we have press conference data available. We choose press conference data because it is available immediately after the meeting as opposed to meeting minutes which are released after at least 21 days. 

\section{Conclusion}
Our work contributes a new cleaned, tokenized, and labeled open-source dataset for FOMC text analysis of various data categories (meeting minutes, speeches, and press conferences). We also propose a new sequence classification task to classify sentences into different monetary policy stances (hawkish, dovish, and neutral). We show the application of this task by generating a measure from the trained model. We validate the measure by studying its relation with CPI, PPI, and Treasury yield. We also propose a simple trading strategy that outperforms the high benchmark set by the QQQ index over the last decade. We release our models, code, and benchmark
data on Hugging Face and GitHub. We also note that the trained model for monetary policy stance classification can be used on other FOMC-related texts. 

\section*{Limitations}
In this article, we focus only on meeting minutes, speech, and press conference data. Many other text datasets such as transcripts from congressional and senate testimonies, beige books, green books, etc can be incorporated to understand pre-FOMC drift better. We don't use audio or video features in constructing the measure, which might contain additional information. It can be an interesting future study to compare measures generated from FOMC text with an alternate measure that can be constructed from the news or social media data. In dataset construction, while splitting sentences, we use a simple rule-based approach. We leave it as an open problem for future researchers to find better methods for splitting sentences with opposite tones. 

In our trading strategy construction, we do not include transaction fees as it involves low-frequency trading. In the future, one can use our model and data to construct a high-frequency trading strategy as well. In addition, a more comprehensive zero-shot and few-shot generative LLM benchmark with open-source models can be performed to provide a better comparison. 

\section*{Ethics Statement}
We acknowledge the geographic bias in our study as we only study the data from the Federal Reserve Bank of the United States of America. We also recognize the presence of gender bias in our study, given the Fed had a female chair for only 4 years out of 27 years (actually the only female chair in its entire history) of the observation period. Data used in the study which will be made public doesn't pose any ethical concerns as all the raw data is public and Fed is subject to public scrutiny. All of the language models used are publicly available and under the license category that allows us to use them for our purpose. Given the pre-training of large PLMs has a big carbon footprint, we limit our work to fine-tuning the existing PLMs. 

\section*{Acknowledgements}
We appreciate the generous infrastructure support provided by Georgia Tech's Office of Information Technology, especially Robert Griffin. We would like to thank Pratvi Shah, Alexander Liu, Ryan Valuyev, and Suraj Chatrathi for their help. We greatly appreciate all the feedback from the reviewers which has helped us improve the paper and add some additional information for readers.

\bibliography{anthology,custom}
\bibliographystyle{acl_natbib}

\appendix
\section{FinBERT Sentiment Analysis}
\label{ap:finbert_senti}
In order to objectively understand the necessity of the new task and the created dataset, we use the \href{https://huggingface.co/ipuneetrathore/bert-base-cased-finetuned-finBERT}{fine-tuned model} available on Hugging-face. The model is fine-tuned for financial sentiment analysis using the pre-trained FinBERT \citep{orig_finbert}. We associate the "positive" label of FinBERT with "dovish", "negative" label with "hawkish", and "neutral" with "neutral" to measure the zero-shot performance on our dataset. The results in Table \ref{tb:finbert_senti} show that the model doesn't perform well, thus reemphasizing the need for a new dataset and task for hawkish-dovish classification. 

\begin{table}[h]
\centering
\footnotesize
\begin{tabular}{lcc}
\hline
\textbf{Data} & \textbf{Mean} & \textbf{Standard Deviation} \\
\hline
MM & 0.3214 & 0.0060 \\
MM-S & 0.3868 & 0.0192 \\
PC & 0.3035 & 0.0253\\ 
PC-S & 0.4357 & 0.0271 \\
SP & 0.5098 & 0.0186 \\
SP-S & 0.5014 & 0.0396 \\
Combined & 0.4254 & 0.0025 \\
Combined-S & 0.4304 & 0.0198 \\
\hline
\end{tabular}
\caption{Here MM indicates that the annotated dataset on meeting minutes is used for training and testing hawkish vs dovish tasks. Similarly, PC stands for press conference data, SP stands for speech data, and Combined is combined data of MM, SP, and PC. *-S indicates the version of the dataset after splitting sentences and reannotation. All  values are F1 scores. 3 seeds were used for all datasets.}
\label{tb:finbert_senti}
\end{table}

\section{Transfer Learning}
\label{ap:transfer_learning}
To understand if there is a need to annotate all three categories of data or whether the model trained on two categories of data can do equally well on the third category, we run an additional experiment. Here we take our best-performing (RoBERTa-large) model and train it on the train split of meeting minutes and press conference combined data and test it on a test sample of speech data. We additionally perform a grid search on four different learning rates (1e-4, 1e-5, 1e-6, 1e-7) and four different batch sizes (32, 16, 8, 4) to find the best hyperparameters. The best average F1 score for 3 seeds is 0.6625 which is lower compared to 0.7169 for the model trained on a training sample of speech data. 

\section{Manual Annotation}
\subsection{Annotation Agreement}
\label{sec:agreement_ann}
Annotation agreement statistics for the split categories of the dataset are provided in Table \ref{tb:annotation_agreement}. Any disagreement between the two annotators was resolved using the annotation guide. If the annotation guide did not cover a specific case of disagreement, online resources were used and the missing case was later added to the annotation guide. 

\begin{table}
\centering
\footnotesize
\begin{tabular}{lc}
\hline
\textbf{Data} & \textbf{Agreement} \\
\hline
MM-S & 89.04\% \\
PC-S & 95.03\% \\
SP-S & 91.13 \\
Combined-S & 90.68 \\
\hline
\end{tabular}
\caption{Annotation agreement statistics for various categories of datasets. }
\label{tb:annotation_agreement}
\end{table}

\subsection{Annotation Guide}
\begin{table*}
\caption{Annotation Guide}
\begin{tabularx}{\textwidth}{c*{3}{>{\raggedright\arraybackslash}X}}
\toprule
 Category & Dovish & Hawkish 
  & Neutral \\
\midrule
Economic Status & when inflation decreases, when unemployment increases, when economic growth is projected as low  & when inflation increases, when unemployment decreases when economic growth is projected high when economic output is higher than potential supply/actual output when economic slack falls
  & When unemployment rate or growth is unchanged, maintained, or sustained \\ 
\midrule
Dollar Value Change & when the dollar appreciates & when the dollar depreciates
  & N/A \\ 
\midrule
Energy/House Prices & when oil/energy prices decrease, when house prices decrease & when oil/energy prices increase, when house prices increase
  & N/A \\ 
\midrule
Foreign Nations & when the US trade deficit decreases & when the US trade deficit increases
  & when relating to a foreign nation's economic or trade policy \\ 
\midrule
Fed Expectations/Actions/Assets & Fed expects subpar inflation, Fed expecting disinflation, narrowing spreads of treasury bonds, decreases in treasury security yields, and reduction of bank reserves & Fed expects high inflation, widening spreads of treasury bonds, increase in treasury security yields, increase in TIPS value, increase bank reserves
  & N/A \\ 
\midrule
Money Supply & money supply is low, M2 increases, increased demand for loans & money supply is high, increased demand for goods, low demand for loans
  & N/A \\ 
\midrule
Key Words/Phrases & when the stance is "accommodative", indicating a focus on “maximum employment” and “price stability” & indicating a focus on “price stability” and “sustained growth”
  & use of phrases “mixed”, “moderate”, “reaffirmed”  \\ 
\midrule
Labor & when productivity increases & when productivity decreases & N/A

\end{tabularx}

\end{table*}
Our annotation guide was built by dividing each target sentence into eight defined categories: Economic Status, Dollar Value Change, Energy/House Prices, Foreign Nations, Fed Expectations/Actions/Assets, Money Supply.
\begin{itemize}
    \item \emph{Economic Status}: A sentence pertaining to the state of the economy, relating to unemployment and inflation
    \item \emph{Dollar Value Change}: A sentence pertaining to changes such as appreciation or depreciation of value of the United States Dollar on the Foreign Exchange Market
    \item \emph{Energy/House Prices}: A sentence pertaining to changes in prices of real estate, energy commodities, or energy sector as a whole.
    \item \emph{Foreign Nations}: A sentence pertaining to trade relations between the United States and a foreign country. If not discussing United States we label neutral.
    \item \emph{Fed Expectations/Actions/Assets}: A sentence that discusses changes in the Fed yields, bond value, reserves, or any other financial asset value. 
    \item \emph{Money Supply}: A sentence that overtly discusses impact to the money supply or changes in demand.
    \item \emph{Key Words/Phrases}: A sentence that contains key word or phrase that would classify it squarely into one of the three label classes, based upon its frequent usage and meaning among particular label classes.
    \item \emph{Labor}: A sentence that relates to changes in labor productivity

\end{itemize}

A label of "Dovish", "Hawkish", and "Neutral" were assigned based on the contents of each sentence by category. The annotation guide and categories were influenced by initial readings of FOMC text and the need to maintain a consistent labeling standard. The annotation guide was utilized during the labeling procedure by two independent annotators to classify each sentence.

Both annotators were male researchers, who have taken finance-related coursework and understood macroeconomics. One originated from the United States, while the other was from India.

\label{sec:appendix_manual_ann}

\section{Robustness check}
\label{ap:robust}
As our dataset is a temporal dataset and the RoBERTa model is trained on data available prior to mid-2019, our model could have utilized future knowledge to predict past sentences a phenomenon deemed "look-ahead bias". Our train-test split based on different seeds contains this bias, so to ensure that it is not present in our model performance, we perform a robustness check by generating a train-test split based on time and checking the performance of the best-performing (RoBERTa-large) model. We split the Combined-S data into a training set spanning from 1996 to 2019 and a test set from 2020 to 2022. For the experiment, we averaged our model performance across 3 seeds (5768, 78516, 944601) and generated an average weighted F1 score of 0.7114, thus validating our performance as not being driven by look-ahead bias. 

\end{document}